\documentclass[conf]{new-aiaa}
\usepackage[utf8]{inputenc}
\usepackage{comment}

\usepackage{graphicx}
\usepackage{amsmath}
\usepackage[version=4]{mhchem}
\usepackage{siunitx}
\usepackage{longtable,tabularx}
\usepackage{booktabs}     
\usepackage{algorithm}
\usepackage{algpseudocode}
\usepackage{float}        
\usepackage{subcaption}
\usepackage{caption}
\usepackage{fixfoot}
\DeclareFixedFootnote{\ieav}{Researcher, C4ISR Division, Institute for Advanced Studies (IEAv).}

\usepackage{hyperref}
\hypersetup{
    colorlinks=true,
    citecolor=red,
    linkcolor=orange,
    filecolor=magenta,      
    urlcolor=orange,
    pdfpagemode=FullScreen,
}

\title{HARBOR: Heading Analysis and Reconstruction from Behavioral Observation and Radar}

\author{%
Joao P. A. Dantas\ieav{}, %
Paulo F. Silva Filho\ieav{}, %
Jelton A. Cunha\ieav{}, %
and %
Gabriel Dietzsch\ieav{}%
}

\affil{Institute for Advanced Studies, São José dos Campos, São Paulo, Brazil, 12.228-001}

\begin{document}

\maketitle

\begin{abstract}
Maritime situational awareness often relies on Automatic Identification System (AIS) transmissions to track vessel movements. However, in operational or conflict scenarios, these data may be unavailable due to signal loss, deliberate deactivation, or intentional spoofing. In such conditions, synthetic aperture radar (SAR) imagery becomes a critical sensing alternative for wide-area maritime monitoring, despite providing only static scene snapshots. This work introduces HARBOR (Heading Analysis and Reconstruction from Behavioral Observation and Radar), a complete pipeline for transforming a single SAR image into predictive motion information without requiring any auxiliary data source at inference time. The method begins with SAR image preprocessing to enhance and segment vessel candidates, followed by automatic detection, size-based classification, and heading estimation using skeleton geometry and local intensity patterns. AIS data are used exclusively during an offline calibration phase to derive vessel-type-dependent motion parameters, which are then applied to generate probabilistic heatmaps of candidate future vessel positions. A case study using real COSMO-SkyMed SAR imagery demonstrates the pipeline on a maritime scene in southern Brazil, showing its ability to extract motion tendencies and generate probabilistic projections of vessel positions in data-denied environments.
\end{abstract}

\section{Introduction}

\lettrine{S}{ynthetic} aperture radar (SAR) imagery is widely used in maritime surveillance, offering all-weather, day-and-night observation capabilities~\cite{martino2020maritime}. These characteristics make SAR particularly valuable for detecting and monitoring vessels across large coastal or open-sea regions, where persistent and reliable coverage is often difficult to achieve using optical sensors alone~\cite{sto2019moving}.

Global maritime traffic has grown substantially in recent decades, encompassing commercial shipping, fishing fleets, and an increasing number of non-cooperative or unregistered vessels engaged in activities such as illegal fishing, smuggling, or unauthorized transit through restricted zones~\cite{galdelli2021sensors}. Monitoring this diverse and dynamic maritime environment demands robust surveillance capabilities that can operate continuously and independently of vessel cooperation~\cite{kanjir2018vessel}.

Maritime situational awareness traditionally relies on Automatic Identification System (AIS) transmissions, which provide vessel identity, position, and motion data derived from onboard GPS sensors~\cite{galdelli2021sensors}. However, in operational or conflict scenarios, AIS data may become unavailable due to equipment failure, intentional deactivation, or deliberate spoofing~\cite{balduzzi2014acsac}. Even under normal conditions, small or uncooperative vessels may not transmit AIS information, creating gaps in the recognized maritime picture. 
Improving situational awareness under incomplete and uncertain information conditions has been a central challenge across defense and maritime operational environments, where data-driven and machine learning approaches have shown strong potential in applications ranging from autonomous navigation and sensor fusion to threat detection and behavior prediction~\cite{thombre2022sensors, nguyen2023sensor, dantas2022machine}. In such data-denied environments, SAR imagery becomes an indispensable sensing alternative for independent vessel detection and behavior inference~\cite{greidanus2017sumo}.

While SAR images provide extensive spatial coverage, they typically capture only a single snapshot of the maritime scene at a given time. Although these images show vessel locations and shapes, they do not directly carry information about motion, intent, or likely future positions. Moreover, the limited revisit frequency of satellite-based SAR restricts the temporal resolution required for continuous tracking~\cite{li2022dlreview}. In time-critical scenarios such as coastal defense, illegal fishing prevention, or maritime traffic monitoring, estimating vessel trajectories and probable movements from a single SAR acquisition can significantly extend situational awareness beyond the moment of capture~\cite{graziano2017wake}.

A fundamental challenge in this problem is that estimating a vessel's trajectory from a single image frame is not straightforward: without temporal information, velocity vectors cannot be directly computed. One class of approaches addresses this by tracking vessels across multiple SAR acquisitions~\cite{renga2011sar}, relying on sufficient revisit frequency to reconstruct motion. Another class combines SAR observations with AIS data at inference time to directly recover velocity and heading information~\cite{heiselberg2023ship}. Both strategies, however, degrade or fail entirely in contested environments where satellite revisit rates are low and cooperative signals are absent or unreliable, precisely the conditions under which robust maritime surveillance is most needed. In such scenarios, simulation-based and data-driven approaches have been increasingly explored to support decision-making under uncertainty, allowing the extraction of behavioral patterns and the prediction of future states from limited or indirect observations~\cite{ribeiro2023ais, yan2022ship, dantas2025simulation}.

This work introduces HARBOR (Heading Analysis and Reconstruction from Behavioral Observation and Radar), a method designed to address this challenge by transforming static SAR imagery into predictive maps of vessel motion. HARBOR avoids the need for temporal sequences or auxiliary data at inference time by separating motion modeling from image analysis: AIS data are used exclusively during an offline calibration phase to derive motion parameters per vessel type, which are then applied directly when processing a new SAR image. The proposed approach performs automatic preprocessing to enhance relevant features and segment potential vessel targets, followed by detection, classification, and heading estimation based on geometric and intensity information. Using these motion parameters, HARBOR associates each vessel category with characteristic speed and angular spread values to generate probabilistic heatmaps of future positions.

The main contributions of this work are:
\begin{itemize}
    \item A fully automated pipeline for vessel detection and heading estimation from a single SAR image, combining morphological segmentation with a bow/stern direction method based on local radar return intensity near the vessel wake;
    \item An offline statistical calibration procedure that extracts motion parameters per vessel type from large-scale AIS datasets, removing the need for cooperative signals at inference time;
    \item A probabilistic heatmap model for short-term trajectory projection, based on calibrated speed and angular spread statistics, that produces direction-aware position estimates from a single SAR image in data-denied environments.
\end{itemize}

The remainder of this paper is organized as follows. Section~\ref{sec:related} reviews related work in SAR-based vessel tracking and trajectory estimation. Section~\ref{sec:methodology} details the HARBOR methodology, while Section~\ref{sec:case_study} presents a case study demonstrating the pipeline application, discussing its outputs and current limitations. Finally, Section~\ref{sec:conclusion} concludes the paper and outlines directions for future work.

\section{Related Work}
\label{sec:related}

Ship detection in SAR imagery has been studied for decades. Early work established automatic detection systems combining land masking, ship candidate search, and wake-based false alarm reduction~\cite{eldhuset1996automatic}. From these foundations, the Constant False Alarm Rate (CFAR) framework became the standard approach, adjusting detection thresholds based on local noise levels to keep false alarm rates under control~\cite{crisp2004state}. Although effective as a baseline, CFAR tends to struggle in dense traffic areas or regions with strong noise sources. Deep learning methods have since substantially improved detection performance, with CNN architectures applied to large-scale SAR datasets achieving notable gains in recall and precision~\cite{li2022dlreview}. Wavelet-based CNNs have also been applied to vessel monitoring in Sentinel-1 imagery, showing that data-driven methods can adapt well to different imaging conditions~\cite{tiwari2021automatized}.

Estimating vessel heading and speed from SAR has also received considerable attention. A common approach applies the Radon or Hough transforms to detect wake features as straight lines in the image, from which heading and speed can be estimated~\cite{graziano2016wake}. These methods work well for large ships with clear wake signatures but lose accuracy for smaller vessels or when the sea surface is rough and wake patterns are hard to distinguish. More recent work uses multitask deep learning to jointly estimate velocity and heading from SAR image data, achieving good results when compared against AIS reference data~\cite{heiselberg2023ship}. These methods, however, still depend on visible wake or Doppler shift signatures and are sensitive to image resolution and sea conditions.

Combining SAR with AIS data has proven useful for improving maritime monitoring. Early studies showed that matching SAR detections with AIS reports helps confirm vessel identities and build a more complete picture of sea traffic~\cite{renga2011sar}. More recent work has gone further by using AIS-based vessel classification to guide the interpretation of SAR detections, making it possible to identify vessel types more reliably from satellite data~\cite{rodger2021classification}. The combination of both data sources has also been applied to fisheries monitoring and the detection of suspicious behavior at sea~\cite{galdelli2021sensors}. All of these approaches, however, require AIS to be available during operation, which is not guaranteed when vessels are uncooperative or in contested areas.

Trajectory forecasting has advanced significantly with sequence-based machine learning. Hybrid models combining Graph Attention Networks with Long Short-Term Memory networks have shown strong performance in predicting vessel movements~\cite{li2024vessel}. More broadly, data-driven methods applied to AIS records have been used to detect anomalous behavior, classify vessel types, and extract motion patterns from historical traffic data~\cite{ribeiro2023ais, yan2022ship}. These methods, however, are trained and applied on historical AIS records and are not designed to work from a single SAR image without prior motion data.

A closely related prior work proposed projecting vessel positions using Monte Carlo simulations applied to SAR detections, assuming vessels had already been identified and classified, but without estimating their heading direction~\cite{falqueto2022persistencia}. HARBOR extends this idea with a complete pipeline covering preprocessing, automatic detection, size classification, and morphological heading estimation, while using AIS data only during an offline calibration step to build motion parameters per vessel type. This design removes the need for any external data source during operation.

SAR image analysis can also draw on richer signal representations beyond intensity. In polarimetric SAR, features based on entropy, anisotropy, and the alpha angle (derived from the statistical decomposition of the polarimetric response) have been used to classify scenes by their dominant signal properties~\cite{cloude1997entropy}. HARBOR currently uses single-polarization intensity data, but these signal-based descriptors suggest a natural direction for improving vessel characterization when fully polarimetric acquisitions are available. Taken together, the methods reviewed here highlight the progress made in SAR-based maritime analysis while also revealing a gap that remains open: no existing approach combines automatic vessel detection, heading estimation, and probabilistic trajectory projection from a single SAR image without requiring auxiliary data at inference time. HARBOR is designed to fill this gap.

\section{Methodology}
\label{sec:methodology}

The HARBOR method consists of a sequence of steps to transform a static SAR image into a predictive map of future vessel positions. The process is automated and uses only information extracted from a single SAR snapshot, without requiring temporal sequences or auxiliary sources. The methodology is divided into five main components: preprocessing, vessel detection, size classification, motion estimation, and trajectory projection. At inference time, the SAR image provides vessel location, extent, and heading information, while future displacement is modeled using AIS-derived motion parameters computed through an offline calibration procedure described in Section~\ref{subsec:trajectory_projection}.

\subsection{Preprocessing}
\label{subsec:preprocessing}

The SAR preprocessing pipeline was implemented using ESA's SNAP toolbox via its Python interface (snappy). The workflow begins with radiometric calibration, which converts raw SAR measurements to sigma-nought ($\sigma^{\circ}$), a standard procedure documented in the Sentinel-1 Product Specification and ESA technical guides~\cite{sentinel1spec,richards2009sar}.

A multilooking operation is then applied to reduce speckle fluctuations and improve the overall image quality of the scene. After multilooking, speckle noise is further reduced using the Lee Sigma filter, an adaptive speckle reduction method introduced by Lee~\cite{lee1980digital}. This filter suppresses image noise while preserving compact high-intensity structures typically associated with vessels.

Next, the image is converted from slant range to ground range through terrain correction using the SRTM 3~arc-second DEM~\cite{farr2007srtm}. Terrain correction compensates for geometric distortions inherent to SAR imaging geometry and ensures that the output image is correctly positioned on the map using standard geographic coordinates~\cite{richards2009sar}. An optional subset operation may then be applied when a region of interest is defined, reducing computational load by cropping the image to the area under analysis.

The final output of the preprocessing chain is written to a GeoTIFF file. This preprocessed image is then normalized to the $[0, 1]$ range and thresholded at $0.99$ to isolate the brightest radar return responses. Morphological opening with a $3 \times 3$ structuring element removes small noise artifacts, followed by morphological closing with a $15 \times 15$ structuring element to reconnect fragmented targets. Connected-component labeling then isolates candidate objects for the vessel detection stage.

\subsection{Vessel Detection}
\label{subsec:vessel_detection}

After preprocessing the SAR image, the next step is to identify regions that may correspond to vessels. This is done by applying connected-component labeling to the binary mask using 8-connectivity, which groups adjacent bright pixels, including diagonal neighbors, into distinct regions. Each connected component receives a unique label so it can be analyzed independently. The choice of 8-connectivity over 4-connectivity ensures that vessel signatures fragmented by minor gaps or irregular shapes are captured as single objects rather than split into multiple components.

For each labeled region, basic geometric properties are computed, including area and bounding box. Regions with fewer than 60 pixels are discarded, as they typically correspond to residual noise or isolated bright returns not associated with vessel targets. This minimum area threshold was determined empirically based on the spatial resolution of the SAR image used in this study and may need to be adjusted for sensors with different ground sampling distances.

More sophisticated detection strategies exist in the literature, including CFAR-based methods~\cite{crisp2004state} and deep learning architectures trained on annotated SAR datasets~\cite{li2022dlreview}. The present work adopts a simpler morphological approach for two reasons: it does not require labeled training data, and it keeps the pipeline fully self-contained with no dependency on external models or pre-trained weights. This design choice is consistent with the operational goal of HARBOR, which prioritizes autonomy and minimal external requirements.

The bounding box of each remaining region is saved for the next stage, where vessels are classified by size. These regions are also used to create individual binary masks that serve as input for the heading estimation step described in Section~\ref{subsec:motion_estimation}.

\subsection{Size Classification}
\label{subsec:size_classification}

The vessels detected in the previous step are grouped into three categories based on their size: small, medium, and large. This is done by computing the area of each vessel’s bounding box in pixels. Vessels with a bounding-box area below 1{,}000~px$^2$ are classified as \textit{Small}, those between 1{,}000 and 5{,}000~px$^2$ as \textit{Medium}, and those above 5{,}000~px$^2$ as \textit{Large}. Although pixel-based area does not directly correspond to physical dimensions, it provides a practical substitute for size categorization when metric vessel length is unavailable. The offline AIS calibration step (Section~\ref{subsec:trajectory_projection}) uses the physical vessel length reported in the AIS record to assign categories under a complementary scheme (Small: $<$~50~m, Medium: 50--200~m, Large: $\geq$~200~m), establishing the mapping between pixel-based size classes and real-world motion parameters.

\subsection{Motion Estimation}
\label{subsec:motion_estimation}

To estimate the direction each vessel is moving, the shape of its binary mask is first reduced to a thin line using skeletonization. This process extracts the central axis of an object while preserving its general shape and structure~\cite{lam1992thinning}.

The two endpoints of the skeleton that are most distant from each other are selected, as they usually represent the front and back of the vessel. An endpoint is defined as a skeleton pixel with exactly one neighboring skeleton pixel in its $3 \times 3$ neighborhood. Then, the mean intensity in a $5 \times 5$ pixel neighborhood around each endpoint is compared. The brighter endpoint is assumed to correspond to the stern, due to the stronger radar return intensity associated with the wake~\cite{crisp2004state}. To quantify the reliability of this assignment, the relative intensity difference between the two endpoints is computed. Vessels for which this difference falls below 10\% are flagged as low-confidence heading estimates, indicating that the bow/stern distinction may be unreliable.

\begin{figure}[!ht]
\centering
\includegraphics[width=0.95\textwidth]{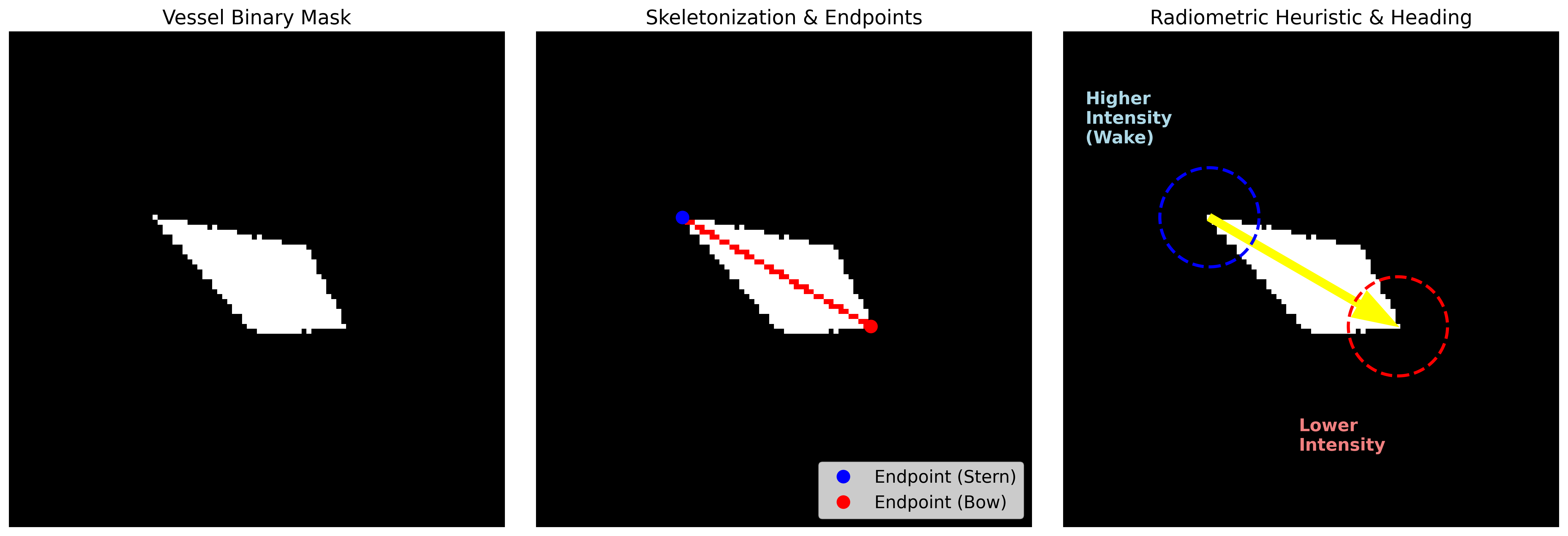}
\caption{Visual representation of the heading estimation method. The binary mask (left) is skeletonized to find the structural endpoints (center). The local radar return intensity around each endpoint is compared to identify the stern (wake) and the bow, defining the motion vector (right).}
\label{fig:method_heading}
\end{figure}

The vessel heading is then estimated as a unit vector pointing from the stern to the bow. This estimated direction is used in the next step to generate the predicted vessel trajectory. Figure~\ref{fig:method_heading} illustrates this heading estimation method.

\subsection{Trajectory Projection}
\label{subsec:trajectory_projection}

After estimating the direction of motion for each vessel, the method predicts its likely future positions by generating a probabilistic heatmap oriented along the inferred heading. The motion parameters used in this prediction are derived from empirical statistics extracted from a public AIS dataset~\cite{marinecadastre2022ais} containing approximately 7~million AIS messages recorded on January~1, 2022, from vessels operating along the entire USA mainland coastline (excluding Alaska). The dataset, sourced from MarineCadastre.gov, provides vessel tracking records including position, speed over ground (SOG), course over ground (COG), heading, and vessel length.

\subsubsection{AIS Data Processing}

The calibration pipeline begins by loading the raw AIS records and applying a sequence of cleaning and filtering steps. First, the compressed AIS fields are decoded: latitude and longitude are divided by $10^5$, SOG by $10$, and COG and heading by $10$, following the standard AIS encoding conventions. Records with invalid or unavailable values are removed, including heading values exceeding $511.0$\textdegree, SOG above $102.3$~knots, and COG outside the $[0, 360)$\textdegree~range. Additionally, records with missing identifiers, timestamps, or coordinates outside valid geographic bounds are discarded.

After cleaning, vessels are classified into three size categories based on the physical length reported in the AIS record: \textit{Small} (length $<$ 50~m), \textit{Medium} (50~m $\leq$ length $<$ 200~m), and \textit{Large} (length $\geq$ 200~m). For each unique vessel (identified by its MMSI), the representative length is determined as the median of all reported length values across its AIS records. Vessels with missing or non-positive length values are excluded from the analysis.

\subsubsection{Statistical Calibration}

For each vessel with at least five AIS data points, two statistics are computed. First, the median SOG is calculated after filtering out records from vessels that were nearly stopped (SOG $\leq$ 0.5~knots), so that only underway speeds contribute to the estimate. Second, the angular dispersion is computed as the standard deviation of consecutive COG differences, capturing how much the vessel deviates from a straight course during its trajectory. The shortest signed angular difference is used to handle the $0$\textdegree/$360$\textdegree~wraparound correctly.

These per-vessel statistics are then aggregated by size category. The representative speed for each category is obtained as the median of all per-vessel median speeds, while the angular dispersion is taken as the median of all per-vessel COG standard deviations. A total of 3{,}301 vessel trajectories were used in the calibration: 2{,}291 small, 623 medium, and 387 large vessels. The resulting parameters are summarized in Table~\ref{tab:parameters}.

\subsubsection{Probabilistic Heatmap Generation}

Using these calibrated parameters, HARBOR generates a two-dimensional Gaussian-like distribution aligned with each vessel's estimated heading. The vessel speed in knots is first converted to pixel displacement per minute using a scale factor specific to the SAR image resolution. The maximum projection radius $r_{\max}$ is then computed as the product of this pixel speed and the desired time horizon.

For each pixel within $r_{\max}$, the angular deviation $\Delta\theta$ from the estimated heading is computed. Only pixels within the directional cone $|\Delta\theta| \leq \sigma_{\theta}/2$, where $\sigma_{\theta}$ is the calibrated angular dispersion, are considered. The probability at each pixel is the product of two Gaussian terms:
\begin{equation}
P(d, \Delta\theta) = \exp\!\left(-\frac{d^2}{2\,(r_{\max}/2)^2}\right) \cdot \exp\!\left(-\frac{\Delta\theta^2}{2\,(\sigma_{\theta}/3)^2}\right)
\end{equation}
where $d$ is the distance from the vessel position and $\sigma_{\theta}$ is the angular dispersion. Both $\Delta\theta$ and $\sigma_{\theta}$ are expressed in radians in Equation~(1); the values reported in Table~\ref{tab:parameters} are converted from the degrees used in the AIS-derived statistics. The distance term uses $\sigma_d = r_{\max}/2$, so that the maximum projection radius corresponds to $2\sigma_d$, capturing approximately 95\% of the probability mass following the standard $2\sigma$ rule of the Gaussian distribution. The angular term uses $\sigma_a = \sigma_{\theta}/3$, so that the directional cone of total width $\sigma_{\theta}$ spans $\pm 1.5\sigma_a$ around the heading direction, capturing approximately 87\% of the angular probability mass. This produces a gradual radial decay along the heading direction and concentrates the probability around the central axis.

\begin{figure}[!ht]
\centering
\includegraphics[width=0.95\textwidth]{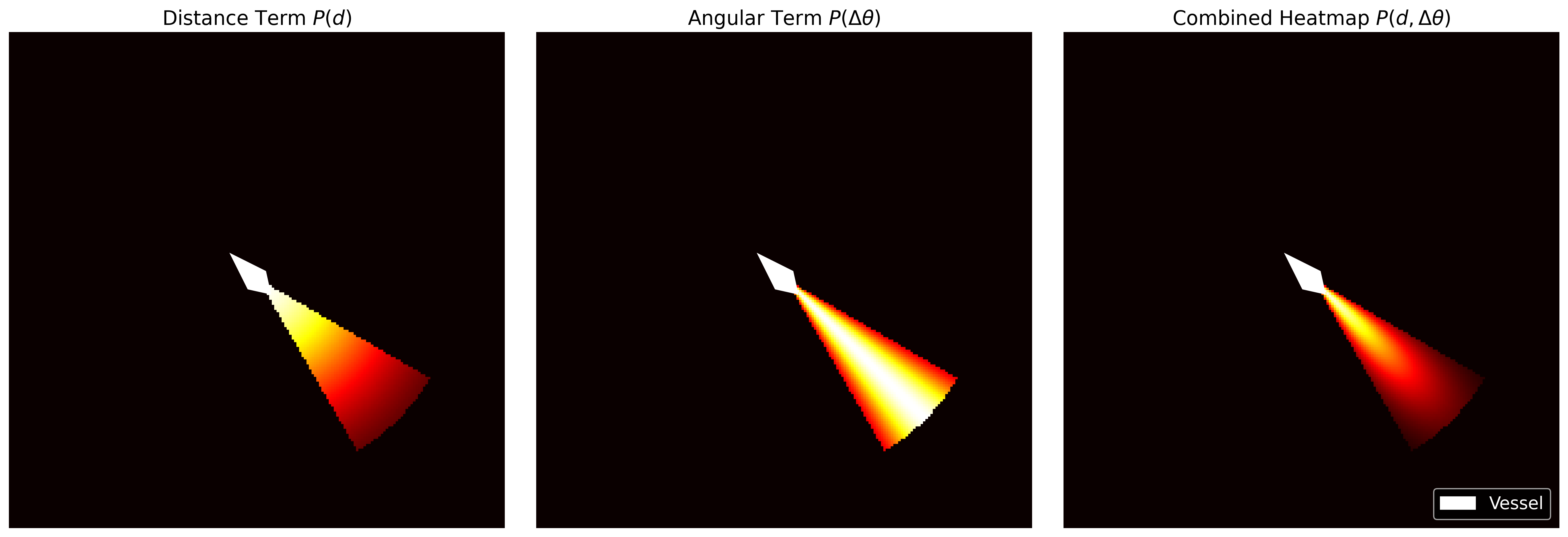}
\caption{Components of the probabilistic motion model. The final local heatmap (right) is the product of a radial distance decay term (left) and an angular deviation cone (center), parameterized by the AIS-derived speed and spread parameters.}
\label{fig:method_heatmap}
\end{figure}

A local heatmap is produced for each detected vessel using this probabilistic model and is placed on the SAR image at the vessel's current position. All local probability fields are then added together to produce a global heatmap, highlighting the regions with the highest expected likelihood of vessel presence over the specified time window. Because individual probability fields are added rather than averaged, areas with several nearby vessels reach higher values, showing the higher chance of finding vessels in places with more traffic. This is a deliberate choice, since the global heatmap is meant to show where vessels are likely to be as a whole, not the probability for each vessel separately. The effect of the spatial and angular Gaussian terms on the resulting probability distribution is shown in Figure~\ref{fig:method_heatmap}.

\begin{table}[H]
\centering
\caption{Motion parameters calibrated from AIS trajectories.}
\label{tab:parameters}
\begin{tabular}{lccc}
\toprule
\textbf{Vessel Category} & \textbf{Median Speed (knots)} & \textbf{Angular Dispersion (degrees)} & \textbf{Sample Count} \\
\midrule
Small   & 4.50  & 31.09 & 2{,}291 \\
Medium  & 10.00 & 5.31  & 623     \\
Large   & 9.60  & 2.35  & 387     \\
\bottomrule
\end{tabular}
\end{table}

\subsection{Implementation Parameters}

Table~\ref{tab:impl_params} summarizes the fixed parameters used in the HARBOR pipeline. These values were determined empirically based on the characteristics of the SAR image used in the case study and may require adjustment for different sensors or acquisition modes.

\begin{table}[H]
\centering
\caption{Fixed parameters of the HARBOR pipeline.}
\label{tab:impl_params}
\begin{tabular}{lcc}
\toprule
\textbf{Parameter} & \textbf{Value} & \textbf{Stage} \\
\midrule
Binarization threshold & 0.99 & Preprocessing \\
Morphological opening kernel & $3 \times 3$ & Preprocessing \\
Morphological closing kernel & $15 \times 15$ & Preprocessing \\
Minimum region area & 60~px & Detection \\
Small vessel area limit & $<$~1{,}000~px$^2$ & Classification \\
Medium vessel area limit & 1{,}000--5{,}000~px$^2$ & Classification \\
Large vessel area limit & $>$~5{,}000~px$^2$ & Classification \\
Intensity neighborhood & $5 \times 5$ & Motion Estimation \\
Minimum AIS points per vessel & 5 & AIS Calibration \\
Minimum SOG threshold & 0.5~knots & AIS Calibration \\
\bottomrule
\end{tabular}
\end{table}

\subsection{Overview}

Algorithm~\ref{alg:HARBOR} summarizes the main steps of the HARBOR method, which processes a single SAR image to detect vessels and estimate their probable future positions through a projected heatmap.

\begin{algorithm}[ht]
\caption{Vessel Trajectory Estimation from a Single SAR Image}
\label{alg:HARBOR}
\begin{algorithmic}[1]
\Require SAR image $I$
\Ensure Projected heatmap $H$
\State Normalize $I$
\State Segment vessel candidates using thresholding and morphological operations
\State Identify connected components $\{R_i\}$
\For{each region $R_i$}
    \If{area$(R_i)$ is below threshold}
        \State Discard $R_i$ as noise
    \Else
        \State Classify $R_i$ as Small, Medium, or Large
        \State Skeletonize $R_i$ and extract endpoints
        \State Estimate heading from skeleton endpoints and local radar return intensity
        \State Generate local heatmap based on direction and vessel parameters
        \State Add local heatmap to global heatmap $H$
    \EndIf
\EndFor
\State \Return $H$
\end{algorithmic}
\end{algorithm}

It is worth noting that the algorithm operates in two distinct phases: an offline phase and an online phase. The offline phase, described in Section~\ref{subsec:trajectory_projection}, processes historical AIS records to derive representative motion parameters for each vessel size category. These parameters are computed once and remain fixed during operation. The online phase covers all steps shown in Algorithm~\ref{alg:HARBOR}: it takes a single SAR image as input and produces a probabilistic heatmap as output, with no dependency on external data sources. This separation is the key design principle of HARBOR, enabling its use in environments where cooperative signals and repeated satellite passes are unavailable.

\section{Case Study and Discussion}
\label{sec:case_study}

To demonstrate the practical application of the HARBOR method, we applied the full pipeline to a real SAR image collected over a maritime region. The goal was to detect vessels, estimate their heading, and generate projected future positions based solely on information extracted from a single frame.

The case study was conducted using a real COSMO-SkyMed Second Generation (CSG) SAR product acquired by satellite SSAR2 over a coastal maritime region in southern Brazil, near the estuary of Lagoa dos Patos and the Port of Rio Grande. The analyzed image corresponds to a Level-1C GEC\_B product acquired in Stripmap mode, left-looking geometry, and VV polarization. The main acquisition parameters are summarized in Table~\ref{tab:sar_metadata}. These metadata are particularly relevant because vessel appearance, wake visibility, and the resulting heading estimation may vary with acquisition geometry, polarization, and spatial resolution.

\begin{table}[ht]
\caption{Main metadata of the SAR product used in the case study.}
\label{tab:sar_metadata}
\centering
\begin{tabular}{lc}
\toprule
\textbf{Parameter} & \textbf{Value} \\
\midrule
Mission & COSMO-SkyMed Second Generation (CSG) \\
Satellite & SSAR2 \\
Processing level & Level-1C \\
Product type & GEC\_B \\
Acquisition mode & Stripmap \\
Polarization & VV \\
Look side & Left \\
Sensing start (UTC) & 2025-03-12 08:59:40 \\
Near / far look angle & $48.60^\circ$ / $49.74^\circ$ \\
Projection & UTM / WGS84 \\
Scene center & ($-32.154^\circ$, $-52.029^\circ$) \\
\bottomrule
\end{tabular}
\end{table}

\begin{figure}[!ht]
\centering
\begin{subfigure}[b]{0.485\textwidth}
    \centering
    \includegraphics[width=\textwidth]{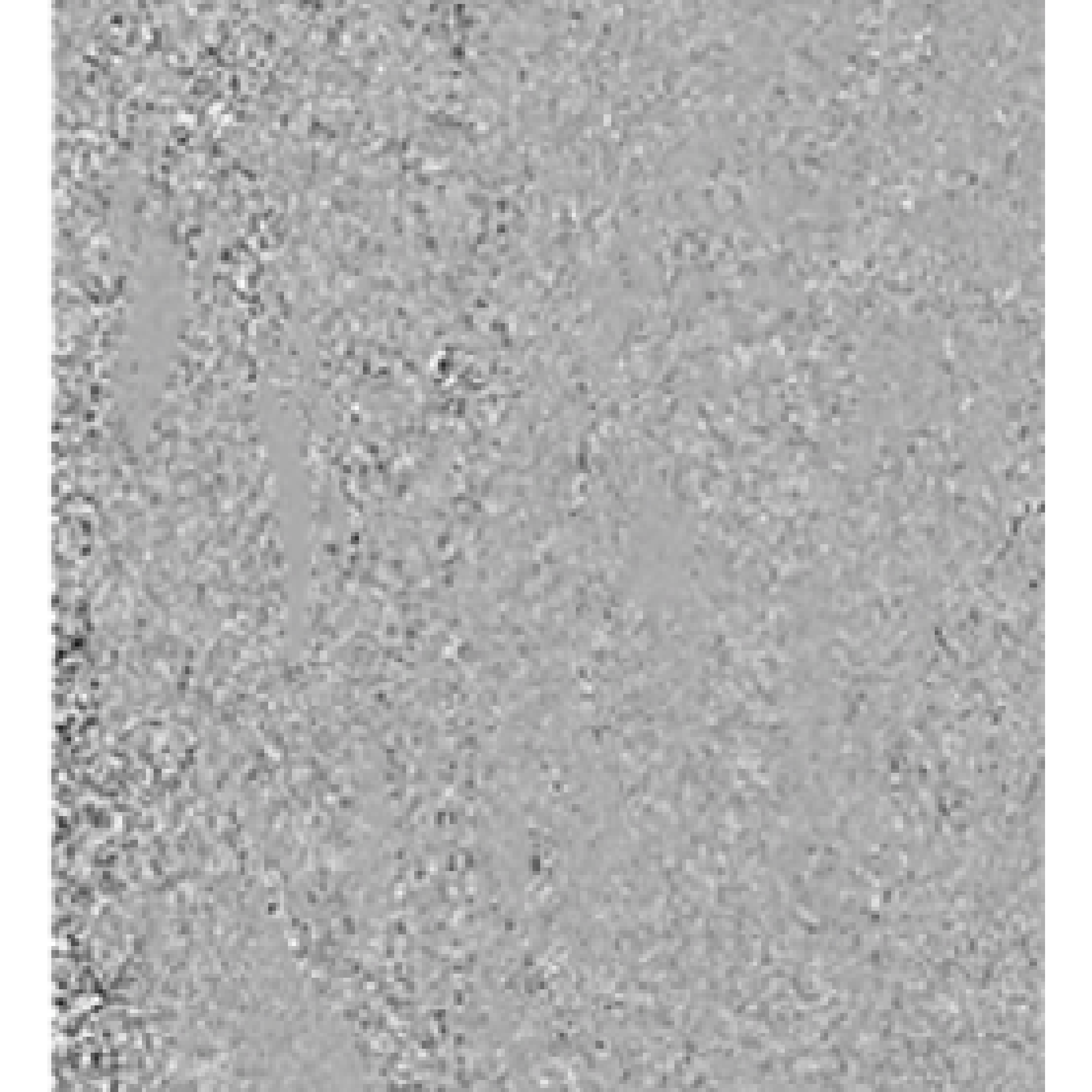}
    \caption{Raw SAR image before preprocessing.}
    \label{fig:raw}
\end{subfigure}
\hfill
\begin{subfigure}[b]{0.485\textwidth}
    \centering
    \includegraphics[width=\textwidth]{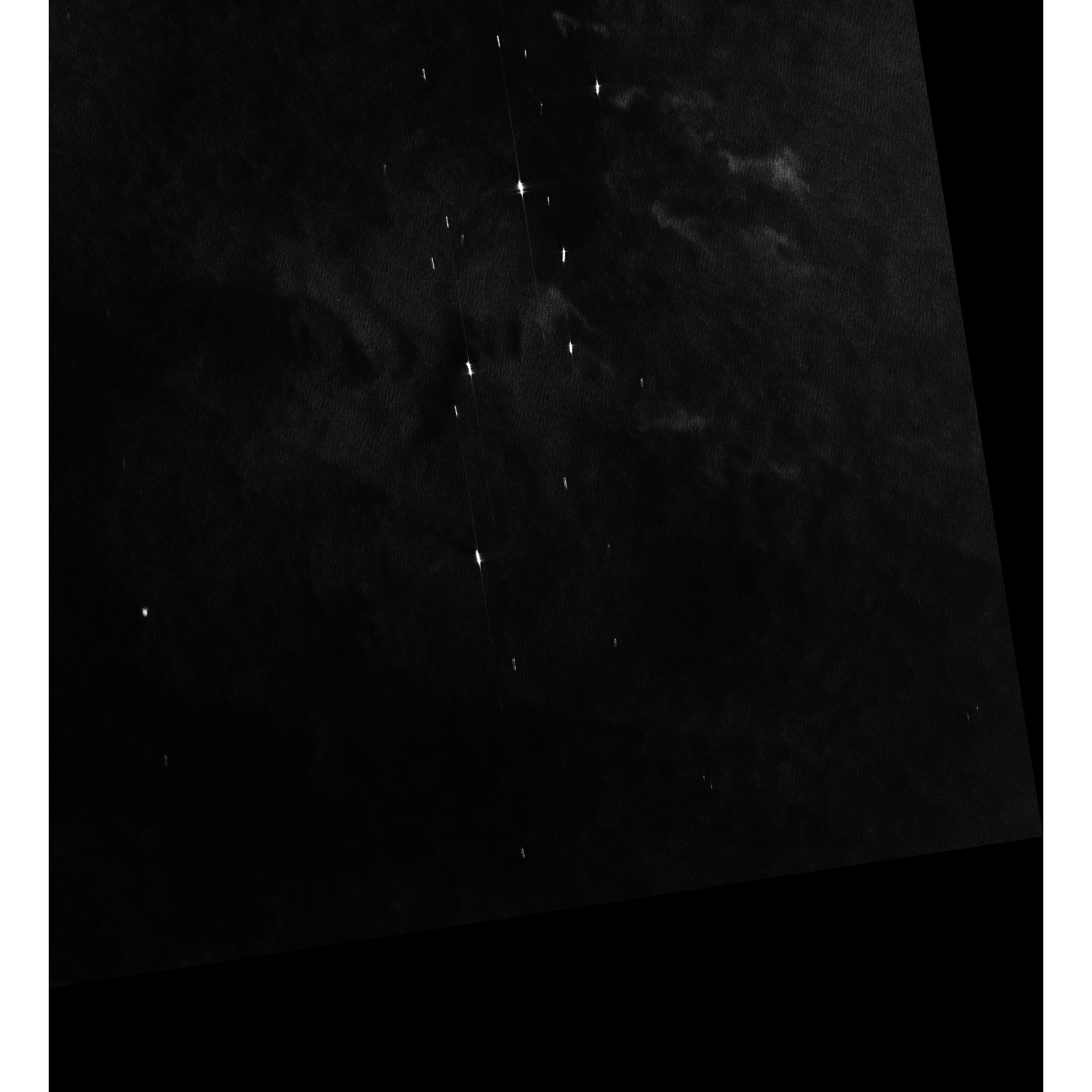}
    \caption{Preprocessed SAR image.}
    \label{fig:preprocessed}
\end{subfigure}

\vspace{0.5em}

\begin{subfigure}[b]{0.485\textwidth}
    \centering
    \includegraphics[width=\textwidth]{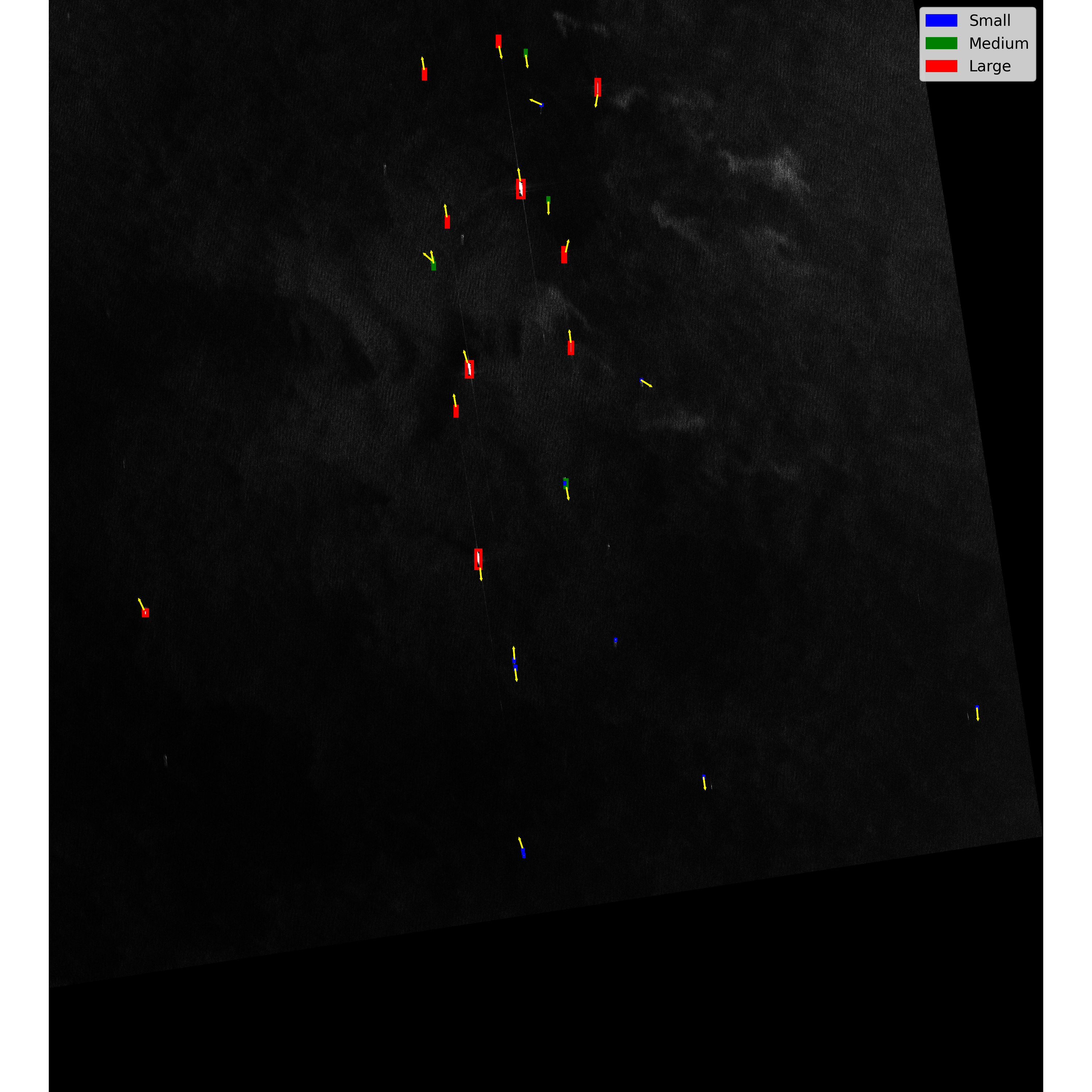}
    \caption{Detected vessels and estimated directions.}
    \label{fig:detections}
\end{subfigure}
\hfill
\begin{subfigure}[b]{0.485\textwidth}
    \centering
    \includegraphics[width=\textwidth]{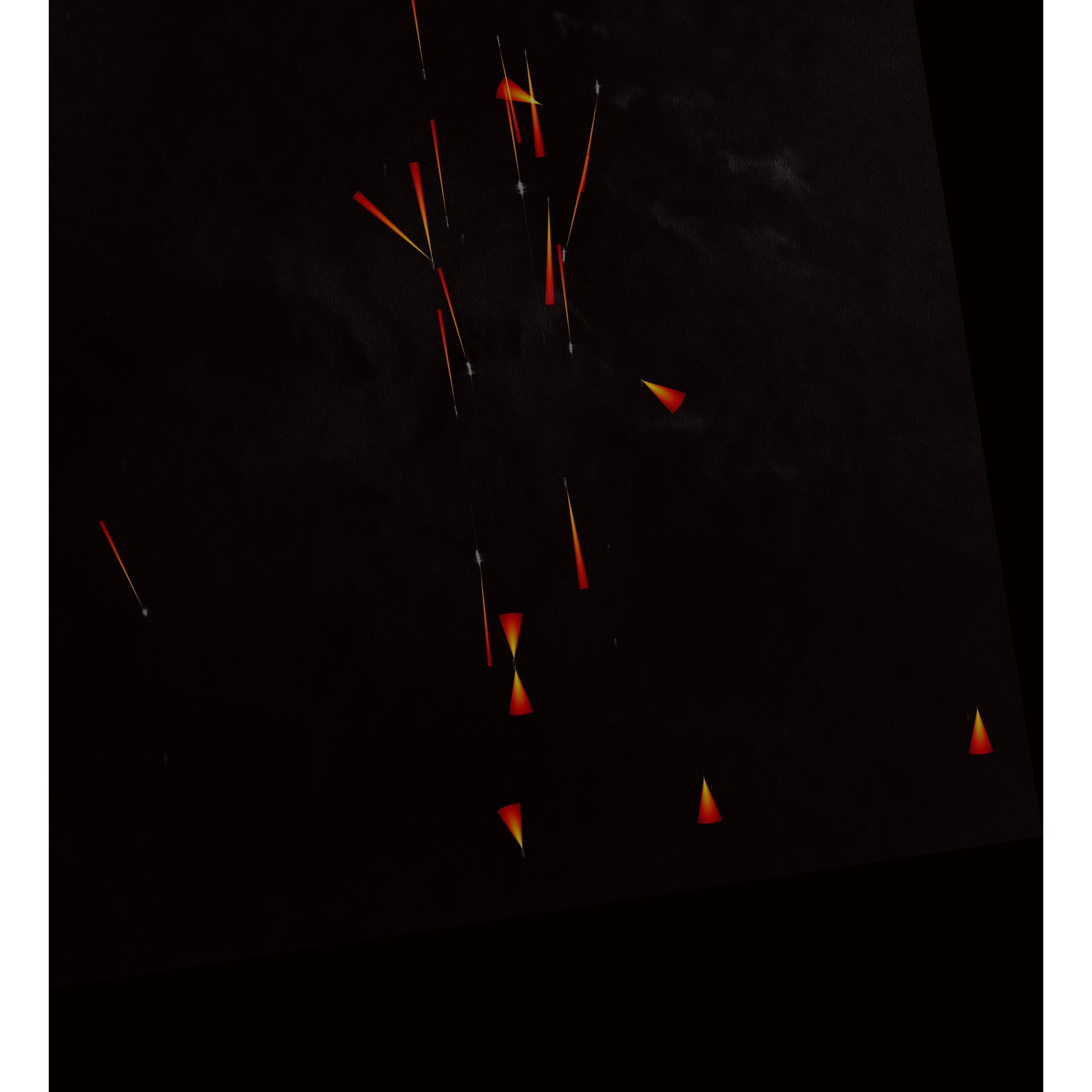}
    \caption{Projected vessel positions for 360 minutes ahead.}
    \label{fig:projection}
\end{subfigure}

\caption{Visual summary of the HARBOR pipeline applied to a COSMO-SkyMed Second Generation SAR image acquired over a coastal maritime region in southern Brazil. (a) Raw SAR image before preprocessing; (b) Preprocessed image after intensity normalization and morphological filtering; (c) Detected vessel candidates with estimated heading directions, color-coded by size category (blue: small, green: medium, red: large); (d) Probabilistic heatmap of projected vessel positions 360~minutes ahead, with brighter regions indicating higher likelihood of vessel presence.}
\label{fig:HARBOR_pipeline}
\end{figure}

The process begins with the raw SAR image, which typically contains a high level of noise and background clutter, making direct interpretation difficult (Figure~\ref{fig:HARBOR_pipeline}a). This input is then passed through the preprocessing stage, where radiometric normalization, thresholding, and morphological operations are applied. These operations enhance vessel-like targets while reducing background noise, resulting in clearer segmented structures (Figure~\ref{fig:HARBOR_pipeline}b).

Next, HARBOR identifies vessel candidates, classifies them into small, medium, or large categories, and estimates their direction of motion. In the evaluated $19{,}217 \times 17{,}496$ pixel scene, the connected-component step initially labeled 58 objects. Filtering by minimum area isolated 27 valid vessel candidates. The processing then analyzed the skeleton of each segmented region and compared local intensity patterns around the endpoints to infer heading. The output includes bounding boxes and directional arrows overlaid on the original image (Figure~\ref{fig:HARBOR_pipeline}c).

Finally, using the estimated heading and AIS-calibrated motion parameters for each vessel size category, the method generates a probabilistic heatmap. The calibrated parameters used in this study were median speeds of 4.50, 10.00, and 9.60~knots, and angular dispersions of $31.09^\circ$, $5.31^\circ$, and $2.35^\circ$ for small, medium, and large vessels, respectively, as summarized in Table~\ref{tab:parameters}. These parameters are used to project the most likely future vessel positions over the selected prediction horizon. A square-root transform is applied to the normalized heatmap before display to enhance the visibility of low-probability regions (Figure~\ref{fig:HARBOR_pipeline}d). The full inference pipeline, including preprocessing and the generation of a 360-minute projection heatmap, completed in 6~minutes and 34~seconds on the evaluated workstation. This processing time is well within the revisit interval of satellite-based SAR systems, which typically ranges from several hours to a few days, confirming the operational viability of the approach.

The 27 detected vessel candidates were distributed across three size categories: 11 small, 5 medium, and 11 large vessels. The relatively high count of large vessels is consistent with the geographic context of the evaluated scene, as the Port of Rio Grande is one of the busiest ports in southern Brazil and a major hub for bulk cargo, container shipping, and offshore support vessels. The presence of small vessels is also expected in this region, given the active artisanal fishing fleet operating in the adjacent coastal waters. The medium category showed the lowest count, which may reflect the tendency of maritime traffic in this area to concentrate at the extremes of the size range, with few vessels of intermediate size. These observations suggest that the size distribution produced by HARBOR is qualitatively consistent with the expected vessel composition for the analyzed scene, even in the absence of reference AIS data for direct verification.

It is important to acknowledge certain limitations of the current HARBOR implementation. The size classification relies on pixel-based area thresholds, assuming a fixed mapping to the physical vessel dimensions derived from AIS records. This mapping is inherently dependent on the spatial resolution and acquisition characteristics of the input SAR image and may therefore require recalibration for data from different sensors or acquisition modes. Additionally, the current implementation relies exclusively on single-polarization intensity data and vessel shape information. When multi-polarization SAR data are available, polarimetric descriptors associated with signal response patterns may provide complementary information for target discrimination and scene interpretation. The wake-based heading method also assumes that the stern produces a stronger radar return. While this assumption is often reasonable for vessels underway, it may fail for slow-moving or anchored ships, or under adverse sea-state conditions that obscure wake signatures. To quantify the reliability of this assignment, the pipeline computes the relative intensity difference between the two skeleton endpoints for each detected vessel. In the evaluated scene, 16 of the 23 vessels with valid heading estimates (69.6\%) exhibited a relative intensity difference below 10\%, indicating low confidence in the bow/stern distinction. This high proportion is consistent with the geographic context of the scene, as the Port of Rio Grande concentrates a large number of stationary or slow-maneuvering vessels for which wake signatures are weak or absent. Therefore, the present case study should be interpreted as a qualitative demonstration of the HARBOR pipeline rather than as a quantitative validation of heading or trajectory accuracy.

\begin{figure}[!ht]
\centering
\includegraphics[width=\textwidth]{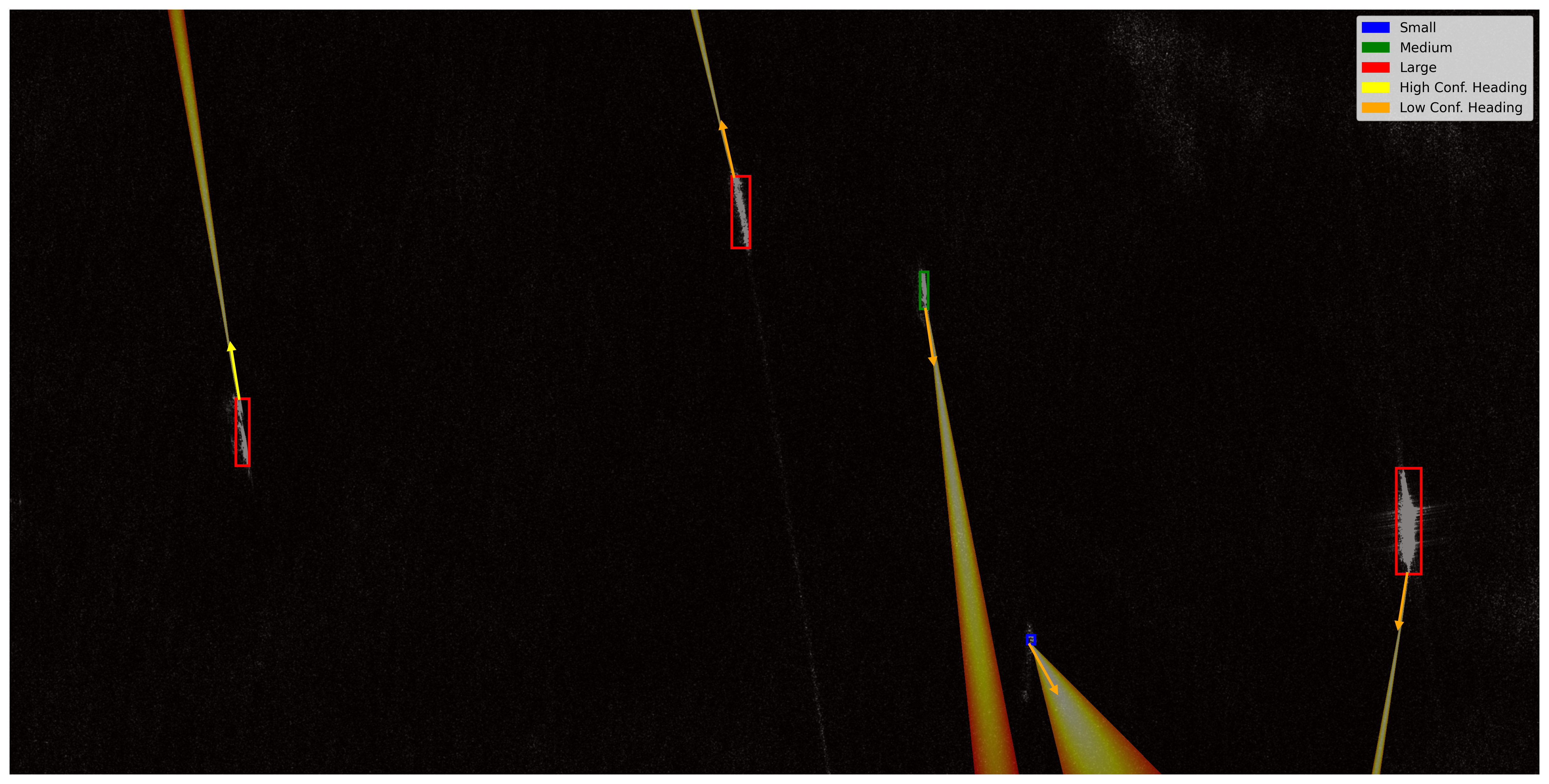}
\caption{Detailed view of a sub-region from the analyzed SAR scene. The image overlays vessel bounding boxes (blue: small, green: medium, red: large), local projected heatmaps, and heading arrows. Yellow arrows indicate high-confidence headings, while orange arrows denote low-confidence estimates (relative intensity difference below 10\%).}
\label{fig:zoom_region}
\end{figure}

To provide a clearer view of the pipeline's outputs at the individual vessel level, Figure~\ref{fig:zoom_region} shows a detailed sub-region of the scene. This composite image overlays the bounding boxes, estimated heading vectors, and local projection heatmaps on the normalized SAR image, illustrating both high- and low-confidence heading assignments across different vessel size categories.

These results suggest that HARBOR can extract motion tendencies from a single SAR snapshot and generate short-term probabilistic projections of vessel positions in data-denied maritime surveillance scenarios.

\section{Conclusion and Future Work}
\label{sec:conclusion}

This paper presented HARBOR, a method for vessel detection and short-term trajectory projection using only a single SAR image at inference time. The approach combines a preprocessing pipeline, automatic vessel detection, size-based grouping, heading estimation from shape and intensity information, and a probabilistic motion model whose parameters are obtained offline from AIS data. By combining single-snapshot SAR inference with AIS-derived motion statistics, the system produces probability maps that show likely vessel movement within a short time window.

The case study using real SAR imagery suggests that HARBOR can capture motion trends and produce direction-aware short-term projections without relying on time sequences or cooperative signals. This shows the method's potential use for maritime surveillance scenarios with limited sensing opportunities.

Several directions are identified for future work. First, and most importantly, a quantitative check using reference AIS tracks is needed to assess heading and trajectory accuracy. This check is essential to measure the method's performance under different sea and traffic conditions, and to find the operational settings where the wake-based heading rule works best. Second, environmental factors such as ocean currents and wind will be added to improve the motion model. Third, the motion model could be extended beyond the three current size categories by using extra image-derived features such as vessel shape, elongation, radar signature strength, or wake structure, allowing more detailed and realistic predictions. Fourth, adding polarimetric SAR descriptors, such as entropy- and alpha-based signal decomposition features, may give richer vessel characterization when multi-polarization data are available. Finally, the AIS records used for calibration came from a single day along the United States coastline, while the case study was carried out in southern Brazil. This gap in location and time adds uncertainty to the calibrated values, and future work will use larger and more varied AIS datasets covering different regions, seasons, and longer time periods.

\section*{Source Code}

The source code developed for this work is publicly available. The preprocessing routines for SAR data are available at \url{https://github.com/jpadantas/sar-preprocessing-snappy}. The code and datasets used for vessel motion modeling and future position prediction are available at \url{https://github.com/jpadantas/harbor}.

\section*{Acknowledgments}

This work was supported by the C2 Project — Command and Control from Massive Aerospace Data Fusion in High-Performance Computing Environments (Agreement No. 01/IEAv/2020), funded by the Department of Aerospace Science and Technology (DCTA), Brazilian Ministry of Defense.

\bibliography{references}

\end{document}